\documentclass[a4paper]{esannV2}
\usepackage[dvips]{graphicx}
\usepackage[latin1]{inputenc}
\usepackage{amssymb,amsmath,array}
\usepackage{todonotes}
\usepackage{multirow}
\usepackage{tikz}
\usepackage{arydshln}
\usepackage{graphicx}
\usepackage{url}

\newcommand{\betadd}{\textit{Beta Distribution Drift Detection}}
\newcommand{\bdd}{\ensuremath{\textit{BD}\textsuperscript{3}}}

\newcommand{\ddm}{\textit{DDM}}
\newcommand{\eddm}{\textit{EDDM}}

\begin{document}
\title{Beta Distribution Drift Detection \\for Adaptive Classifiers}

\author{Lukas Fleckenstein, Sebastian Kauschke, and Johannes F\"urnkranz
%
\vspace{.3cm}\\
%
TU Darmstadt -- Knowledge Engineering Group \\
Hochschulstrasse 10, 64289 Darmstadt, Germany
%
}

\maketitle

\begin{abstract} 
With today's abundant streams of data, the only constant we can rely on is change. For stream classification algorithms, it is necessary to adapt to concept drift.
This can be achieved by monitoring the model error, and triggering counter measures as changes occur.
In this paper, we propose a drift detection mechanism that fits a beta distribution to the model error, and treats abnormal behavior as drift. It works with any given model, leverages prior knowledge about this model, and allows to set application-specific confidence thresholds. Experiments confirm that it performs well, in particular when drift occurs abruptly.
\end{abstract}

\section{Introduction}

In the domain of learning from data streams, it is highly probable that the target concept will change over time, i.e, that concept drift will affect the stream. This usually leads to the necessity of updating the model that deals with the data. Hence, many learning algorithms specifically geared towards this situation have been investigated. Some of them deal with drift implicitly, while others require explicit drift detection. A popular explicit approach towards the supervised learning scenario is to monitor the performance of the learner via its accuracy or error rate. When a significant change in the model's performance is detected, a mechanism for coping with this problem is invoked, for example by updating the model with current information \cite{Bifet2007,Gama2004,baena2006}.

In this paper, we present \betadd{} (\bdd), a method that leverages previously known information about an existing classifier in the form of a beta distribution, and detects drift by assessing if new batches of data operate within the confidence bounds of that distribution. If previous knowledge does not exist, we gather it in the beginning of the drift detection process. The method works on batches of data, as opposed to other methods that work on single instances at a time. The batchwise approach is generally more stable w.r.t.\ the trade-off between false alerts and false negatives, and, as we think, applies to more real-world scenarios.

In section~\ref{sec:problem} we define the problem, followed by the explanation of our algorithm in section~\ref{sec:bddd}. We describe the experimental setup in section~\ref{sec:experiment}, discuss our results in section~\ref{sec:results}, and draw some conclusions in section~\ref{sec:conclusion}.

\section{Problem Definition}
\label{sec:problem}
We assume a model $M$ which classifies a stream of data instances $D$. The stream consists of batches $D_i$, $i \in 0...I$, where the number of batches $I$ is large or potentially infinite. 
For each batch $D_i$, the model $M$ classifies the $n_i$ instances, producing a corresponding error batch $E_i$.
Each error batch $E_i$ consists of binary classifier errors $e_{i,j}$, where $e_{i,j} = 0$ if $M$ predicts instance $j \in 0...n_i$ correctly, and $e_{i,j}=1$ otherwise. 
The goal is to detect whether a batch $E_{i}$ shows a significant increase in error rate compared to previous batches. We will approach this by fitting beta distributions to the classifier error, as described in the next section.

\section{Beta Distribution Drift Detection Method}
\label{sec:bddd}
The method presented in this paper is based on evaluating the beta distribution of the classification error of a model. For each new batch of data, the current classifier error     is compared against this distribution. If it is outside a certain confidence interval, concept drift is assumed.

\paragraph{Model Initialization.}
The beta distribution has two shape parameters $\alpha, \beta > 0$, which we initially derive from previous knowledge about the error-rate $\pi_0$ of the model. As a rule of thumb we use $\alpha_0 = \pi_0 \cdot n_0$ and $\beta_0 = (1 - \pi_0) \cdot n_0$, where $n_0$ is the number of instances in the first batch that we receive.
If $\pi_0$ is unknown, we suggest 
to set it to $0.5$ as a starting value.

\paragraph{Batchwise Model Update.}
Given the most recent batch $D_i$, the detector receives the corresponding binary error batch $E_i$. For a given error rate $p_i$ of the model on this batch, the likelihood for observing $k_i$ misclassifications on the $n_i$ samples is 
given by the binomial distribution $\textit{Bin}(k_i|n_i, \pi_i)$.
By putting a prior on that error rate, we are able to compute the posterior probability of the error rate given some data. Since it is a conjugate prior to the binomial, we choose a beta distribution  $\textit{Beta}(\pi_i|\alpha_i, \beta_i)$, where $\alpha_i$ and $\beta_i$ represent the number of previously misclassified and correctly classified samples, respectively.
It can be shown that: 
\begin{equation}
P(\pi{}_i|E_i) =  \frac{P(E_i|\pi{}_i) \cdot P(\pi{}_i)}{P(E_i)} = \frac{\textit{Bin}(k_i|n_i, \pi{}_i) \cdot \textit{Beta}(\pi{}_i|\alpha_i, \beta_i)}{P(E_i)} = \textit{Beta}(\alpha_i^*, \beta_i^*)
\end{equation}
with $\alpha_i^* = \alpha_i + k_i \label{equ:pseudocounts_a}$
and
$\beta_i^* = \beta_i + (n_i-k_i) \label{equ:pseudocounts_b}$.

\begin{figure}[!ht]
    \centering
    \includegraphics[width=\textwidth]{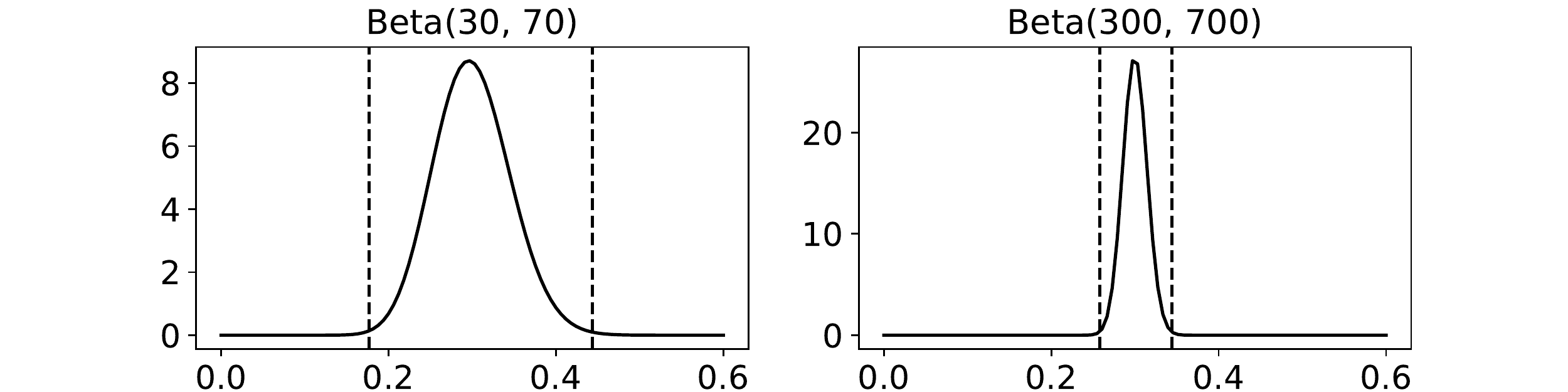}
    \caption{Two example beta probability density functions with their boundaries that contain 99.7\% of the distribution.}
    \label{fig:betadistribution}
\end{figure}

\paragraph{Drift Detection.}
We can now 
test if the error of a new batch of data is likely to correspond to the classifier's concept, or if a concept change occurred:
\begin{enumerate}
	\item Compute warn and drift boundaries that contain 95.0\% and 99.7\% of the distribution $\textit{Beta}(\alpha^*_{i-1}, \beta^*_{i-1})$, and the current error rate $\pi_i = k_i / n_i$.\\
	            If $\pi_i > upper\_bound_{warn}$ signal warning. \\
	            If  $\pi_i > upper\_bound_{drift}$ signal drift and reset shape parameters to \\ 
	            $\alpha^*_{i-1} = \alpha_0 \ , \ \beta^*_{i-1} = \beta_0$. Reset test counter $t = 0$. 
	\item Set the parameters of the previous posterior distribution $\textit{Beta}(\alpha^*_{i-1}, \beta^*_{i-1})$ as prior for the most recent one, $\alpha_i = \alpha^*_{i-1}, \ \beta_i = \beta^*_{i-1}$.  
	\item Compute the most recent shape parameters \\
	        $\alpha_i^* = \frac{\alpha_i}{\text{decay}} + k_i$, $\beta_i^* = \frac{\beta_i}{\text{decay}} + (n_i - k_i)$. Increment test counter $t = t + 1$.
\end{enumerate}
The parameter $decay$ is used to control the influence of prior knowledge given by the previous batches. As the reliability of that prior increases with the number of tests $t$, we decrease $decay$ according to $decay = 1 / \exp(a \cdot(t + b)) + 1.1$. We choose $a=0.15, \ b=-7$ based on preliminary experiments as a trade-off between abrupt and gradual drift detection.
It can be shown that $\lim_{i\to\infty} \alpha^*_i + \beta^*_i = n + \frac{n}{decay-1} \ \ \forall \ decay > 1$.
Thus, the decay parameter limits the variance of the beta distribution, preventing it from becoming too narrow, which would increase false alerts.
Figure~\ref{fig:betadistribution} shows how different shape parameters $\alpha$ and $\beta$ influence the beta distribution, even if their mean $\pi = \alpha / (\alpha + \beta)$ remains the same. 

\section{Experiment Setup}
\label{sec:experiment}
In this section, we describe the datasets we chose for evaluation, the evaluation procedure, the used metrics, and the algorithms we compare against.

\subsection{Benchmark Algorithms}
We compare our novel approach against two tried-and-tested approaches which follow the idea of detecting drift from classifier error:\footnote{Implemented in \url{https://scikit-multiflow.github.io}}
\begin{description}
    \item[\ddm] The Drift Detection Method by Gama et al. \cite{Gama2004} relies on detecting statistically significant changes in the performance of a classifier via monitoring the mean error and its standard deviation. A drift is detected if a certain threshold $\sigma$ is exceeded.
    \item[\eddm] The Early Drift Detection Method \cite{baena2006} is similar to \ddm{}. Instead of monitoring the error rate, it monitors the distances between subsequent errors. This allows to better detect gradual and slow changes, thus eliminating a weakness of \ddm{}.
\end{description}

\subsection{Evaluation Datasets}
We evaluate our findings on four datasets. Each scenario represents a different type of concept drift with varying severity and/or gradation:

\begin{description}
\item[Bit-Stream.] The first dataset is a stream of bits from a Bernoulli distribution with parameter $\mu$ as proposed in \cite{FriasBlanco2015}. Each stream contains 30 change points, separated by 600 or 2000 bits for which the concept is stable. In order to simulate different drift magnitudes, the maximum absolute difference between the means of two subsequent concepts is restricted in an interval $[a, b]$. 


\item[SEA Concepts.] SEA is a drifting stream generation scheme \cite{Street_2001} and used as a standard test for abrupt concept change. The dataset has three features, two determine the class and the third is noise. We generate streams of 40k instances with three change points at 10k, 15k, and 30k samples, and also vary between 10\% and 20\% class noise.


\item[Rotating Hyperplane.] This dataset has features equal to SEA, but contains a gradual drift behavior. Class labels depend on the placement of the two-dimensional points compared to a hyperplane that rotates during the course of the stream. It starts rotation with a certain angle every 1,000 instances, starting after the first 10k samples, the angles being 20\textdegree, 30\textdegree, and 40\textdegree.


\item[Elec2.] This dataset \cite{Harries_1999} is a real-world dataset of electricity prices. It contains 45,312 instances with eight features and binary class labels that indicate price change (UP or DOWN), and has an unknown number of drifts.
\end{description}

\subsection{Evaluation Measures}
\label{sec:evaluationmeasures}
For the comparison of the algorithms we use the following metrics:
\begin{description}
	\item[FPR:] The false positive rate, where the drift detection method detects a drift when there is actually none.
	\item[FNR:] The false negative rate, where the model fails to detect a drift, when there is one.
	\item[Delay:] The average number of batches between the true drift point and the first true positive detection on the same concept.
	
\end{description}
We run each algorithm 50 times with a batch size of 200 on every dataset and calculate the average values and standard deviations of FPR, FNR, and Delay for those runs.
For the synthetic and real-world datasets we use a Na\"ive Bayes classifier in the \textit{Interleaved Test-Then-Train} or \textit{Prequential} framework, in which a new batch is first used for evaluating the accuracy and afterwards for updating the classifier (cf.~\cite{Gama_2014} for more details).

\begin{table}[t]
\centering
\caption{Results on the Bit-Stream dataset.}
\label{tab:bstreamabrupt}
\scriptsize
\begin{tabular}{p{1cm}p{0.5cm}p{1.2cm}p{1.2cm}p{1.2cm}p{1.2cm}p{1.2cm}p{1.2cm}}
\hline
\multirow{2}{*}{Algorithm} &       & \multicolumn{3}{c}{600 bits between changes}                                                                                                                                                     & \multicolumn{3}{c}{1,000 bits between changes}                                                                                                                                                   \\ \cline{3-8} 
                           &       & \hspace{4pt}{[}0.1, 0.3{]}                                                 & \hspace{4pt}{[}0.3, 0.5{]}                                                 & \hspace{4pt}{[}0.5, 0.7{]}                                                 & \hspace{4pt}{[}0.1, 0.3{]}                                                 & \hspace{4pt}{[}0.3, 0.5{]}                                                 & \hspace{4pt}{[}0.5, 0.7{]}                                                 \\ \hline
DDM                        & FPR   & \begin{tabular}[c]{@{}c@{}}\textbf{0.0310}\\   ($\pm$ 0.0232)\end{tabular} & \begin{tabular}[c]{@{}c@{}}\textbf{0.0215}  \\ ($\pm$ 0.0133)\end{tabular} & \begin{tabular}[c]{@{}c@{}}\textbf{0.0197}\\   ($\pm$ 0.0127)\end{tabular} & \begin{tabular}[c]{@{}c@{}}\textbf{0.0106}\\   ($\pm$ 0.0071)\end{tabular} & \begin{tabular}[c]{@{}c@{}}\textbf{0.0062}\\   ($\pm$ 0.0050)\end{tabular} & \begin{tabular}[c]{@{}c@{}}\textbf{0.0085}\\   ($\pm$  0.0055)\end{tabular} \\ \cline{2-8} 
                           & FNR   & \begin{tabular}[c]{@{}c@{}}0.5579\\   ($\pm$ 0.1996)\end{tabular} & \begin{tabular}[c]{@{}c@{}}0.2254\\   ($\pm$ 0.1975)\end{tabular} & \begin{tabular}[c]{@{}c@{}}\textbf{0.0115}\\   ($\pm$ 0.0523)\end{tabular} & \begin{tabular}[c]{@{}c@{}}0.5011\\   ($\pm$ 0.2180)\end{tabular} & \begin{tabular}[c]{@{}c@{}}0.2180\\   ($\pm$ 0.2783)\end{tabular} & \begin{tabular}[c]{@{}c@{}}\textbf{0.0114}\\   ($\pm$ 0.0800)\end{tabular} \\ \cline{2-8} 
                           & Delay & \begin{tabular}[c]{@{}c@{}}0.6122\\   ($\pm$ 0.3011)\end{tabular} & \begin{tabular}[c]{@{}c@{}}0.3325\\   ($\pm$ 0.1367)\end{tabular} & \begin{tabular}[c]{@{}c@{}}0.0273\\   ($\pm$ 0.0468)\end{tabular} & \begin{tabular}[c]{@{}c@{}}2.3346\\   ($\pm$ 0.9044)\end{tabular} & \begin{tabular}[c]{@{}c@{}}1.4450\\   ($\pm$ 0.3424)\end{tabular} & \begin{tabular}[c]{@{}c@{}}0.4567\\   ($\pm$  0.1183)\end{tabular} \\ \hline
EDDM                       & FPR   & \begin{tabular}[c]{@{}c@{}}0.1719\\   ($\pm$ 0.0832)\end{tabular} & \begin{tabular}[c]{@{}c@{}}0.1831\\   ($\pm$ 0.0588)\end{tabular} & \begin{tabular}[c]{@{}c@{}}0.2197\\   ($\pm$ 0.0461)\end{tabular} & \begin{tabular}[c]{@{}c@{}}0.0861\\   ($\pm$ 0.0458)\end{tabular} & \begin{tabular}[c]{@{}c@{}}0.1018\\   ($\pm$ 0.0359)\end{tabular} & \begin{tabular}[c]{@{}c@{}}0.1099\\   ($\pm$ 0.0299)\end{tabular} \\ \cline{2-8} 
                           & FNR   & \begin{tabular}[c]{@{}c@{}}0.4072\\   ($\pm$ 0.1942)\end{tabular} & \begin{tabular}[c]{@{}c@{}}0.0923\\   ($\pm$ 0.1138)\end{tabular} & \begin{tabular}[c]{@{}c@{}}\textbf{0.0000}\\   ($\pm$ 0.0000)\end{tabular} & \begin{tabular}[c]{@{}c@{}}0.4656\\   ($\pm$ 0.1731)\end{tabular} & \begin{tabular}[c]{@{}c@{}}0.0950\\   ($\pm$ 0.1438)\end{tabular} & \begin{tabular}[c]{@{}c@{}}\textbf{0.0000}\\   ($\pm$ 0.0000)\end{tabular} \\ \cline{2-8} 
                           & Delay & \begin{tabular}[c]{@{}c@{}}0.3508\\   ($\pm$ 0.2182)\end{tabular} & \begin{tabular}[c]{@{}c@{}}0.4337\\   ($\pm$ 0.1317)\end{tabular} & \begin{tabular}[c]{@{}c@{}}0.0530\\   ($\pm$ 0.0603)\end{tabular} & \begin{tabular}[c]{@{}c@{}}2.0603\\   ($\pm$ 0.8537)\end{tabular} & \begin{tabular}[c]{@{}c@{}}2.5416\\   ($\pm$ 0.4280)\end{tabular} & \begin{tabular}[c]{@{}c@{}}1.3600\\   ($\pm$ 0.2792)\end{tabular} \\ \hline
\bdd              & FPR   & \begin{tabular}[c]{@{}c@{}}0.0521\\   ($\pm$ 0.0377)\end{tabular} & \begin{tabular}[c]{@{}c@{}}0.0944\\   ($\pm$ 0.0402)\end{tabular} & \begin{tabular}[c]{@{}c@{}}0.0618\\   ($\pm$ 0.0406)\end{tabular} & \begin{tabular}[c]{@{}c@{}}0.0552\\   ($\pm$ 0.0335)\end{tabular} & \begin{tabular}[c]{@{}c@{}}0.0802\\   ($\pm$ 0.0422)\end{tabular} & \begin{tabular}[c]{@{}c@{}}0.0383\\   ($\pm$ 0.0286)\end{tabular} \\ \cline{2-8} 
\multicolumn{1}{l}{}       & FNR   & \begin{tabular}[c]{@{}c@{}}\textbf{0.0312}\\   ($\pm$ 0.0414)\end{tabular} & \begin{tabular}[c]{@{}c@{}}\textbf{0.0000}\\   ($\pm$ 0.0000)\end{tabular} & \begin{tabular}[c]{@{}c@{}}\textbf{0.0000}\\   ($\pm$ 0.0000)\end{tabular} & \begin{tabular}[c]{@{}c@{}}\textbf{0.0270}\\   ($\pm$ 0.0421)\end{tabular} & \begin{tabular}[c]{@{}c@{}}\textbf{0.0000}\\   ($\pm$ 0.0000)\end{tabular} & \begin{tabular}[c]{@{}c@{}}\textbf{0.0000}\\   ($\pm$ 0.0000)\end{tabular} \\ \cline{2-8} 
\multicolumn{1}{l}{}       & Delay & \begin{tabular}[c]{@{}c@{}}\textbf{0.0189}\\   ($\pm$ 0.0383)\end{tabular} & \begin{tabular}[c]{@{}c@{}}\textbf{0.0000}\\   ($\pm$ 0.0000)\end{tabular} & \begin{tabular}[c]{@{}c@{}}\textbf{0.0000}\\   ($\pm$ 0.0000)\end{tabular} & \begin{tabular}[c]{@{}c@{}}\textbf{0.0637}\\   ($\pm$ 0.1246)\end{tabular} & \begin{tabular}[c]{@{}c@{}}\textbf{0.0000}\\   ($\pm$ 0.0000)\end{tabular} & \begin{tabular}[c]{@{}c@{}}\textbf{0.0000}\\   ($\pm$ 0.0000)\end{tabular} \\ \hline
\end{tabular}
\end{table}

\section{Results}
\label{sec:results}
\begin{figure}[ht!]
    \centering
    \includegraphics[width= \textwidth]{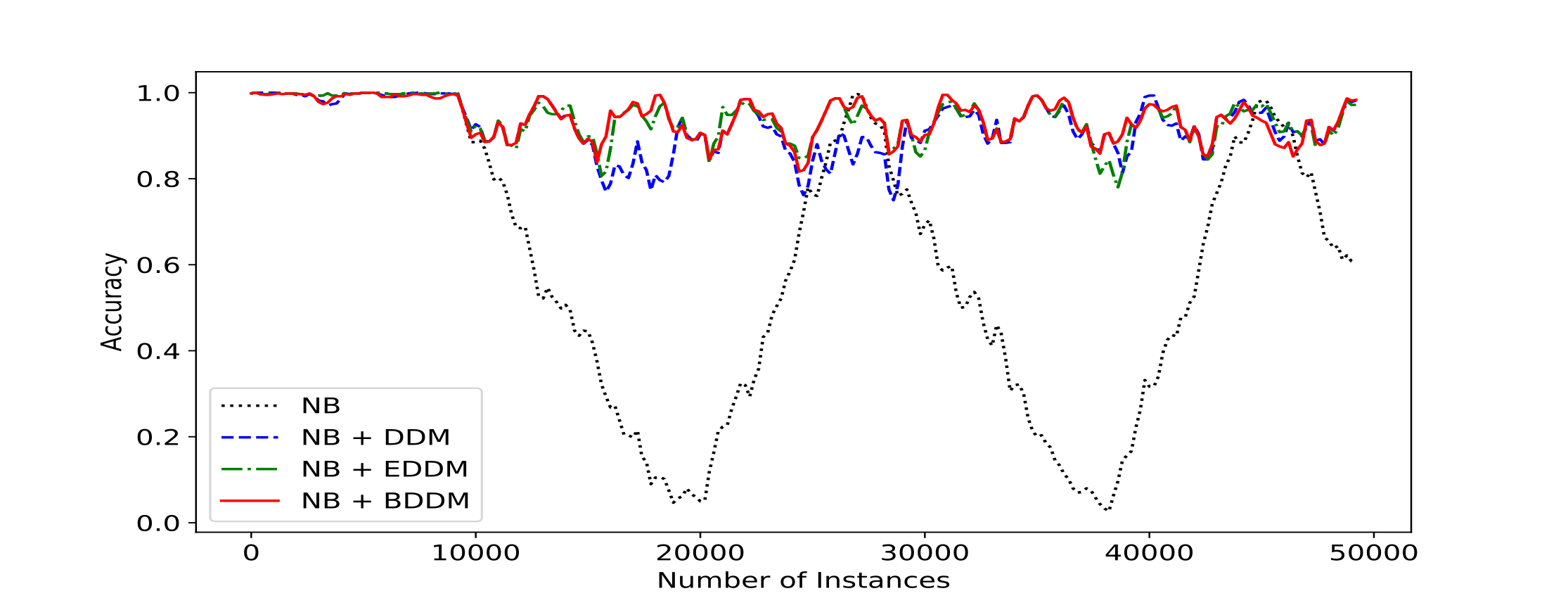}
    \caption{Accuracy comparison on the Rotating Hyperplane dataset (20\textdegree).}
    \label{fig:accuracyresults}
\end{figure}

In Table~\ref{tab:bstreamabrupt},    we show the results on the abruptly drifting bit-stream. Compared to \ddm{} and \eddm{}, \bdd{} generally shows low FNR and Delay values. Especially for the smallest change interval of $[0.1, 0.3]$, where \ddm{} and \eddm{} show high levels of FNR, \bdd{} performs significantly better.
%
%
On the synthetic datasets (Table~\ref{tab:synth}), \bdd{} shows comparable results to \ddm{} and \eddm{}. Here, the performance depends heavily on the classifier algorithm, and the chosen drift detector has limited effect. However, the accuracy is marginally above \ddm{} and \eddm{}, which we attribute to the lower delay achieved by \bdd{}. Figure~\ref{fig:accuracyresults} shows the accuracy as a stream on the gradually drifting rotating hyperplane dataset, where \bdd{}s advantage in delay shows by faster recovery on dips in the curve. 
On the Elec2 dataset (Table~\ref{tab:elec}), \bdd{} is the only drift detector that does not lower the accuracy below the levels of no detector. 

\begin{table}[t]
\centering
\caption{Results on synthetic datasets.}
\label{tab:synth}
\scriptsize
\begin{tabular}{cccc|ccc}
\hline
\multirow{2}{*}{Algorithm}      & \multicolumn{3}{c}{SEA Concepts (Abrupt)}                                                                                                                                                                          & \multicolumn{3}{c}{Rotating Hyperplane (Gradual)}                                                                                                                                                         \\ \cline{2-7} 
                                & 0.0                                                            & 0.1                                                            & 0.2                                                            & 20                                                             & 30                                                             & 40                                                             \\ \hline
\multicolumn{1}{l}{No Detector} & \begin{tabular}[c]{@{}c@{}}0.9137\\ ($\pm$  0.0014)\end{tabular}   & \begin{tabular}[c]{@{}c@{}}0.8487 \\ ($\pm$ 0.0015)\end{tabular}  & \begin{tabular}[c]{@{}c@{}}0.7631  \\ ($\pm$ 0.0015)\end{tabular} & \begin{tabular}[c]{@{}c@{}}0.6046\\   ($\pm$ 0.0079)\end{tabular} & \begin{tabular}[c]{@{}c@{}}0.5933\\   ($\pm$ 0.0106)\end{tabular} & \begin{tabular}[c]{@{}c@{}}0.5834\\   ($\pm$ 0.0080)\end{tabular} \\ \cline{2-7} 
DDM                             & \begin{tabular}[c]{@{}c@{}}0.9417 \\ ($\pm$ 0.0111)\end{tabular}  & \begin{tabular}[c]{@{}c@{}}0.8643\\  ($\pm$ 0.0102)\end{tabular}  & \begin{tabular}[c]{@{}c@{}}0.7671\\   ($\pm$  0.0100)\end{tabular} & \begin{tabular}[c]{@{}c@{}}0.9221\\   ($\pm$ 0.0073)\end{tabular} & \begin{tabular}[c]{@{}c@{}}0.9027\\   ($\pm$ 0.0131)\end{tabular} & \begin{tabular}[c]{@{}c@{}}0.8925\\   ($\pm$ 0.0110)\end{tabular} \\ \cline{2-7} 
EDDM                            & \begin{tabular}[c]{@{}c@{}}0.9471\\  ($\pm$ 0.0023)\end{tabular}  & \begin{tabular}[c]{@{}c@{}}0.8621\\  ($\pm$ 0.0036)\end{tabular}  & \begin{tabular}[c]{@{}c@{}}0.7617\\   ($\pm$ 0.0038)\end{tabular} & \begin{tabular}[c]{@{}c@{}}0.9283\\   ($\pm$ 0.0064)\end{tabular} & \begin{tabular}[c]{@{}c@{}}0.9109\\   ($\pm$ 0.0095)\end{tabular} & \begin{tabular}[c]{@{}c@{}}0.9039\\   ($\pm$ 0.0068)\end{tabular} \\ \cline{2-7} 
\bdd                            & \begin{tabular}[c]{@{}c@{}}\textbf{0.9496}\\   ($\pm$ 0.0024)\end{tabular} & \begin{tabular}[c]{@{}c@{}}\textbf{0.8717}\\   ($\pm$ 0.0039)\end{tabular} & \begin{tabular}[c]{@{}c@{}}\textbf{0.7750}\\   ($\pm$ 0.0057)\end{tabular} & \begin{tabular}[c]{@{}c@{}}\textbf{0.9315}\\   ($\pm$ 0.0061)\end{tabular} & \begin{tabular}[c]{@{}c@{}}\textbf{0.9133}\\   ($\pm$ 0.0085)\end{tabular} & \begin{tabular}[c]{@{}c@{}}\textbf{0.9053}\\   ($\pm$ 0.0058)\end{tabular} \\ \hline
\end{tabular}
\end{table}

\begin{table}[hb]
\centering
\caption{Results on the Elec2 dataset.}
\label{tab:elec}
\scriptsize
\begin{tabular}{lccccccc}
\hline
\multicolumn{1}{c}{} & No Detector & DDM    & EDDM   & \bdd  \\
Accuracy             & 0.7270      & 0.7232 & 0.7243 & \textbf{0.7307}  \\ \hline
\end{tabular}
\end{table}

\section{Conclusion}
\label{sec:conclusion}
In this paper, we have shown a novel drift detection method that monitors the classifier error via a beta distribution. Change in the classifier's performance is detected as drift, the sensitivity of the detector can be adjusted via a confidence threshold and decay parameters. Existing knowledge about the classifier's performance can be used to set the initial parameters of the distribution, which allows immediate drift detection. 
Experimental results show that the method is robust against false positives, while simultaneously being fast in detecting concept drift. 


\begin{footnotesize}


\bibliographystyle{unsrt}
\bibliography{references,kauschke,ensembles,drift}

\end{footnotesize}


\end{document}